%% file: main.tex
\documentclass[10pt,twocolumn,letterpaper]{article}

\usepackage{cvpr}              

\usepackage{adjustbox} 

\input{preamble}

\definecolor{cvprblue}{rgb}{0.21,0.49,0.74}
\usepackage[pagebackref,breaklinks,colorlinks,citecolor=cvprblue]{hyperref}

\def\paperID{14} 
\def\confName{CVPR}
\def\confYear{2024}

\title{Fractals as Pre-training Datasets for Anomaly Detection and Localization}

\author{Cynthia I. Ugwu, Sofia Casarin, Oswald Lanz\\
Free University of Bozen-Bolzano, Bolzano, Italy\\
{\tt\small \{cugwu,scasarin,oswald.lanz\}@unibz.it}
}

\begin{document}
\maketitle
\input{sec/0_abstract}    
\input{sec/1_intro}

\subsection*{Acknowledgements} The research presented in this paper was partially funded by \textit{Covision Lab}, the Italian \textit{National Operative Program}, budget for ”Research and Innovation” 2014-2020 under Action IV.5 ”Doctorates on green topics” and Free University of Bozen-Bolzano startup fund IN2814 and EG2403

{
    \small
    \bibliographystyle{ieeenat_fullname}
    \bibliography{main}
}


\end{document}

%% file: preamble.tex
\usepackage[dvipsnames]{xcolor}


%% file: sec/0_abstract.tex
\begin{abstract}
Anomaly detection is crucial in large-scale industrial manufacturing as it helps detect and localise defective parts. Pre-training feature extractors on large-scale datasets is a popular approach for this task. Stringent data security and privacy regulations and high costs and acquisition time hinder the availability and creation of such large datasets. While recent work in anomaly detection primarily focuses on the development of new methods built on such extractors, the importance of the data used for pre-training has not been studied. Therefore, we evaluated the performance of eight state-of-the-art methods pre-trained using dynamically generated fractal images on the famous benchmark datasets MVTec and VisA.
In contrast to existing literature, which predominantly examines the transfer-learning capabilities of fractals, in this study, we compare models pre-trained with fractal images against those pre-trained with ImageNet, without subsequent fine-tuning. Although pre-training with ImageNet remains a clear winner, the results of fractals are promising considering that the anomaly detection task required features capable of discerning even minor visual variations. This opens up the possibility for a new research direction where feature extractors could be trained on synthetically generated abstract datasets reconciling the ever-increasing demand for data in machine learning while circumventing privacy and security concerns.
\end{abstract}

%% file: sec/1_intro.tex
\section{Introduction}
\label{sec:intro}
Identifying unusual structures in images is a challenging problem in computer vision with numerous applications, including industrial inspection (\cite{bergmann2019mvtec, bergmann2022beyond}), healthcare monitoring (\cite{zimmerer2022mood, menze2014multimodal}), autonomous driving (\cite{blum2019fishyscapes, hendrycks2019scaling}), and video surveillance (\cite{liu2018future, nazare2018pre}). Due to the rarity and complexity of determining the full specification of defect variations, most of the literature addresses the Anomaly Detection (AD) problem unsupervised, where a model is only trained on anomaly-free images. However, obtaining training data is expensive and time-consuming, and privacy concerns limit availability, especially in industrial and medical scenarios. Recently computer vision systems have expanded greatly as large-scale datasets, such as ImageNet, have led to a shift from model-driven to data-driven approaches (\cite{kataoka2020pre}). For example, in AD, many current state-of-the-art methods rely on deep feature extractors pre-trained on a proxy task on large-scale datasets. In addition to the technical challenges and high costs associated with acquiring and labelling these large datasets, questions have arisen over privacy, ownership, inappropriate content, and unfair biases. This has resulted in ImageNet being restricted to non-commercial applications, the 80M Tiny Images dataset (\cite{torralba200880}) being withdrawn, promising datasets such as JFT-300M (\cite{sun2017revisiting}) or Instagram-3.5B (\cite{mahajan2018exploring}) being unavailable for public use, and LAION-5B (\cite{schuhmann2022laion}), which was used to train the famous Stable Diffusion (\cite{rombach2021highresolution}), being withdrawn due to ethical concerns. 

\textit{``What if we had a way to harness the power of large image datasets with few or none of the major issues and concerns currently faced? (\cite{anderson2022improving})"}. Fractals are complex geometric structures generated by mathematical equations, thus, anyone can produce the images making them open-source, without the necessity of massive manual labelling and ethical or bias concerns. The work of \cite{kataoka2020pre} was the first to introduce the possibility of using fractals as an alternative pre-trining method for image recognition tasks. In light of the promising results shown in image classification (\cite{anderson2022improving}) and 3D scene understanding (\cite{yamada2022point}), in this paper, we conduct extensive experiments to examine the potential utility of using a synthetically generated dataset composed of fractals for the detection and localization of industrial anomalies. This study differs from the existing literature that mainly focuses on fractals' transfer-learning (fine-tuning) ability for supervised classification, we compared the AD methods pre-trained with fractals against ImageNet without fine-tuning, introducing additional complexity to the comparison as the model's weights remain untuned, thereby lacking familiarity with the dataset. Moreover, defect detection is a challenging task as normal and abnormal samples look very similar but differ in local appearance, necessitating robust features capable of discerning even minor visual variations, while classification tasks involve semantically distinct classes, simplifying the discrimination process.
Our contributions are summarised as follows:
\begin{itemize}
    \item We conducted the first systematic analysis comparing the performance of AD models pre-trained with fractals against ImageNet.
    \item We analyze the impact of feature hierarchy and object categories in solving the AD task, showing that low-level fractal features are more effective and emphasizing the importance of anomaly type selection when considering fractal images.
    \item Our findings motivate a new research direction in AD, where there is the potentiality to replace large-scale natural datasets with completely synthetic abstract datasets reducing annotation labour, protecting fairness, and preserving privacy.
\end{itemize}

\section{Related Works}
Most unsupervised AD models can be divided into two main groups: (i) reconstruction-based and (ii) feature embedding-based methods. In this paper, we focus on the latter. Feature embedding-based methods rely on the ability to learn the distribution of anomaly-free data by extracting descriptors from a pre-trained backbone (feature extractor) that most of the time is kept frozen during the entire AD process. Anomalies are detected during inference as deviations from these anomaly-free features, assuming the feature extractor produces different features for anomalous images. 
According to \cite{xie2023iad}, feature embedding-based methods can be divided into four categories: teacher-student (\cite{wang2021student, deng2022anomaly, bergmann2020uninformed, guo2023recontrast}), memory bank (\cite{roth2022towards, defard2021PaDiM, lee2022cfa}), normalizing flow (\cite{gudovskiy2022cflow, yu2021fastflow}), and one-class classification (\cite{reiss2021panda, li2021cutpaste}). For teacher-student models, during the training phase, the teacher is the feature extractor and distils the knowledge to the student model. When an abnormal image is passed, the teacher will produce features that the student wasn't trained on, so the student network won't be able to replicate the features. Thus, the feature difference in the teacher-student network is the most important principle in detecting anomalies during inference. Regarding memory bank-based approaches features of normal images are extracted from a pre-trained network and stored in a memory bank. Test samples are classified as anomalous if the distance between the extracted test feature and the closest neighbourhood feature point inside the memory bank exceeds a certain threshold. Normalizing flow is used to learn transformations between data distributions. In AD, anomaly-free features are extracted from a pre-trained network and projected by the trainable flow model to an isotropic Gaussian distribution, in other words, the model applies a change of variable formula to fit an arbitrary density to a tractable base distribution. During inference, the normalizing flow is used to estimate the precise likelihood of a test image. Anomalous images should be out of distribution and have a lower likelihood than normal images. For one-class classification, the goal is to identify instances belonging to a single class, without explicitly defining the boundaries between classes as in traditional binary classification.

\section{Fractals Images}
Fractal images are generated using Iterated Function Systems (IFS), composed of two or more functions, each associated with a sampling probability. Affine IFS involves affine transformations: $\omega(x)= Ax + b$, where $A$ represents a linear function and $b$ represents a translation vector. The set of functions has an associated set of points with a particular geometric structure called \textit{attractor}. 
Following the definition of \cite{anderson2022improving} and \cite{kataoka2020pre}, an IFS system $S$, with cardinality $N \sim U(\{2,3, .., 8\})$, defined on a complete metric space $\mathcal{X} =(\mathrm{I\!R}^2,\|\cdot\|_2)$ is a set of transformations $\omega_i:\mathcal{X}\rightarrow\mathcal{X}$ and their associated probabilities $p_i$:
\begin{equation}
    S = {(\omega_i,p_i): i=1,2...N}
\end{equation}
which satisfy the average contractility condition. The attractor $\mathcal{A}_S$ is a unique geometric structure, a subset of $\mathcal{X}$ defined by $\mathcal{S}$. The shape of $\mathcal{A}_S$ depends on the function $\omega_i$, while the sampling probabilities $p_i \propto |detA_i|$ influence the distribution of points on the attractor that are visited during iterations. Affine transform parameters are associated with the categories of the synthetic dataset. \cite{anderson2022improving} improved the sampling strategy to always guarantee the contractility condition of $S$ and produce fractals with ``good" geometric properties. Fractal images with good geometry are not too sparse, containing complex and varied structures with few empty spaces. An affine transform must have singular values less than 1 to be a contraction, which can be imposed by construction. Thus, the authors used singular values decomposition of $A = U\Sigma V^T$, where $U$ and $V$ are orthogonal matrices and $\Sigma$ is a diagonal matrix containing the singular values $\sigma_1$ and $\sigma_2$. By sampling $\sigma_1$ and $\sigma_2$ in the range (0, 1), we ensure the system is a contraction. 
Regarding good geometry, the authors empirically demonstrate that singular values' magnitudes dictate how quickly an affine contraction map converges to its fixed point under iteration. Small values cause quick collapse, while values near 1 lead to ``wandering" trajectories which don’t converge to a clear geometric structure. They empirically find that given $\sigma_{i,1}$ and $\sigma_{i,2}$ be the singular values for $A_i$, the \textit{i}th function in the system, the majority of the systems with good geometry satisfy $\frac{1}{2}(5+N)<\alpha<\frac{1}{2}(6+N)$ with $\alpha$ being 
\begin{equation}
    \alpha = \sum_{i=1}^{N} (\sigma_{i,1} + 2\sigma_{i,2}).
\end{equation}
For $N=2,...,8$, the founded range works well, although it may also work for a wider range.

\begin{figure*}
    \centering
    \includegraphics[width=0.9\textwidth]{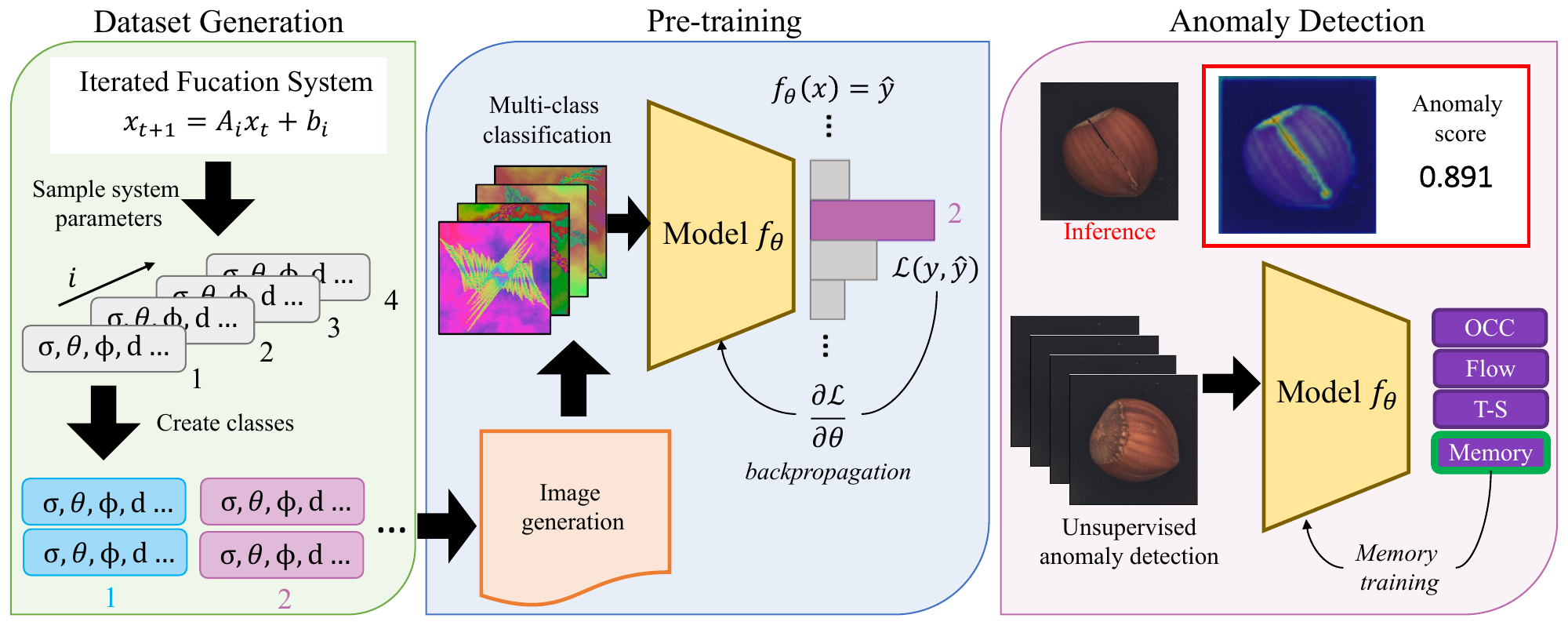}
    \caption{We generate a dataset of IFS codes by sampling the parameters of the system which are used to generate fractals images. The generated images are used to train a computer vision model for multi-class classification. Finally, the model is used as a feature extractor for unsupervised anomaly detection.}
    \label{fig:framework}
\end{figure*}

\section{Implementation Details}
Our synthetic dataset, named ``Fractals" for simplicity, consists of single-fractal images for multi-class classification obtained by grouping 100,000 IFS into 1000 classes. We follow the default configuration of \cite{anderson2022improving}. We trained WideResNet50 with the standard cross-entropy objective function for 100 epochs using 1,000,000 training samples per epoch with an image resolution of $256\times 256$ and batch size of 512. For the anomaly detection, we used teacher-student methods RD (\cite{deng2022anomaly}), STFPM (\cite{wang2021student}), memory-based methods PatchCore (\cite{roth2022towards}), PaDiM (\cite{defard2021PaDiM}), the flow models FastFlow (\cite{yu2021fastflow}), C-Flow (\cite{gudovskiy2022cflow})
and the one-class classification methods PANDA (\cite{reiss2021panda}), and CutPaste (\cite{li2021cutpaste}). 
To facilitate reproducibility, we used Anomalib (\cite{akcay2022anomalib}) to train the anomaly detection methods, except for PANDA and CutPaste deployed through the official code implementations. The overall framework can be seen in \cref{fig:framework}.

\textbf{Datasets:} To study industrial anomaly detection performance, our experiments are performed on the MVTec (\cite{bergmann2019mvtec}) and VisA (\cite{zou2022spot}).
MVTec contains 15 sub-datasets of industrially manufactured objects. For each object class, the test sets contain both normal and abnormal samples with various defect types. The dataset is relatively small scale, where the number of training images for each sub-dataset varies from 60 to 391, posing a unique challenge for learning deep representations. VisA contains 12 sub-datasets. The objects range from different types of printed circuit boards to samples with multiple or single instances in a view.

\textbf{Evaluation Metrics:} Image-level metrics are used to assess AD algorithms' classification performance, whereas pixel-level metrics are used to assess their segmentation (localization) performance. These two types of metrics represent distinct capabilities of AD algorithms, and they are both extremely important. Following prior work we use the area under the receiver operator curve (AUROC) for both image-level and pixel-level anomaly detection. To measure localization performance we also use the area under the per-region-overlap (AUPRO). In contrast to the ROC measure which is biased in favour of large anomalies, the PRO score weights ground-truth regions of different sizes equally to better account for varying anomaly sizes, see (\cite{bergmann2020uninformed}) for details.

\section{Results}
\label{sec:results}
\begin{table*}[htbp]
    \centering\fontsize{7.7}{11}\selectfont
    \begin{tabular}{l|cccccccc}
        \toprule
        Class & \textbf{FastFlow }& \textbf{C-Flow} & \textbf{PatchCore} & \textbf{PaDiM} & \textbf{RD }&\textbf{ STFPM }& \textbf{CutPaste} & \textbf{PANDA} \\
        \midrule
        carpet & 98.6/64.5 & 92.7/49.6 & 98.0/40.9 & {\color{red}99.0}/42.5 & 98.9/30.6 & 98.0/53.9 & 85.9/{\color{RoyalBlue}69.2} & 93.4/31.2 \\
        grid & 99.8/58.0 & 96.1/82.0 & 97.5/93.7 & 96.9/78.5 & {\color{red}100.0}/68.3 & 98.3/46.0 & 98.3/{\color{RoyalBlue}100.0} & 52.0/54.4 \\
        leather & 99.7/{\color{RoyalBlue}88.5} & 96.1/63.3 & {\color{red}100.0}/82.0 & 99.7/81.9 & {\color{red}100.0}/75.2 & 99.8/67.9 & {\color{red}100.0}/87.3 & 96.5/54.4 \\
        tile & 99.9/95.6 & 99.9/92.7 & 98.8/95.6 & 99.5/{\color{RoyalBlue}97.3} & {\color{red}100.0}/60.9 & 98.6/74.0 & 94.7/84.8 & 96.8/65.1 \\
        wood & 99.2/{\color{RoyalBlue}99.6} & 95.6/93.8 & 99.4/{\color{RoyalBlue}97.9} & 99.1/97.1 & 99.4/84.3 & {\color{red}99.7}/75.5 & {\color{red}99.7}/95.7 & 95.9/56.8 \\
        bottle & {\color{red}100.0}/97.6 & {\color{red}100.0}/56.7 & {\color{red}100.0}/88.2 & 99.8/95.9 & 99.9/93.2 & {\color{red}100.0}/54.9 & 99.8/{\color{RoyalBlue}97.9} & 96.8/65.1 \\
        cable & 92.9/55.6 & 92.0/45.9 & {\color{red}98.8}/52.2 & 93.2/61.4 & 96.2/58.6 & 91.3/43.9 & 90.6/{\color{RoyalBlue}85.8} & 84.5/54.9 \\
        capsule & 94.7/42.1 & 90.4/61.6 & {\color{red}97.8}/73.4 & 91.9/70.8 & 97.6/{\color{RoyalBlue}78.3} & 57.9/56.5 & 83.5/78.1 & 91.8/71.8 \\
        hazelnut & 97.9/{\color{RoyalBlue}97.6} & 99.6/85.7 & {\color{red}100.0}/92.0 & 94.1/93.9 &{\color{red} 100.0}/89.5 & {\color{red}100.0}/90.8 & 97.2/71.3 & 88.5/61.3 \\
        metal\_nut & 98.7/57.8 & 96.4/34.4 & 99.8/38.1 & 98.7/47.9 & {\color{red}100.0}/69.8 & 96.6/66.2 & 94.2/{\color{RoyalBlue}80.7} & 72.9/41.5 \\
        pill & 96.4/{\color{RoyalBlue}79.5} & 82.4/76.5 & 93.1/75.9 & 92.3/77.2 & {\color{red}96.7}/72.4 & 81.0/77.4 & 89.1/71.0 & 81.0/65.3 \\
        screw & 85.0/27.5 & 89.1/69.0 & 97.9/61.7 & 85.2/40.0 & {\color{red}98.1}/{\color{RoyalBlue}69.1} & 90.3/60.4 & 79.0/42.75 & 70.5/41.3 \\
        toothbrush & 77.5/60.8 & 71.4/78.3 & {\color{red}100.0}/99.2 & 87.2/98.6 & 93.9/96.7 & 85.0/79.2 & 87.8/{\color{RoyalBlue}97.8} & 88.1/68.9 \\
        transistor & 89.7/59.7 & 87.8/33.0 & {\color{red}99.9}/55.2 & 98.5/78.6 & 97.4/66.8 & 94.9/37.5 & 92.8/{\color{RoyalBlue}79.8} & 91.0/71.2 \\
        zipper & 89.3/74.4 & 91.6/44.6 & 99.3/81.2 & 88.3/76.8 & 98.3/{\color{RoyalBlue}83.2} & 81.5/46.5 & {\color{red}99.8}/70.9 & 97.0/57.6 \\
        \hline
        \textbf{Model Avg }& 94.6/70.6 & 92.1/64.5 & \textbf{{\color{red}98.7}}/75.1 & 94.9/75.9 & 98.4/73.1 & 91.5/62.0 & 92.8/\textbf{{\color{RoyalBlue}80.9}} & 86.4/57.4 \\
        \bottomrule
        \end{tabular}
        \caption{MVTec image-level AUROC. Each cell carries the results for ImageNet/Fractals.}
        \label{tab:mvtec_image_auroc}
\end{table*}

\begin{table}[htbp]
\begin{adjustbox}{width=\columnwidth}
   \centering\fontsize{7.7}{11}\selectfont
  \begin{tabular}{l|cccccccc|}
    \toprule
    \textbf{Class} & \textbf{FastFlow} & \textbf{C-Flow} & \textbf{PatchCore} & \textbf{PaDiM} & \textbf{RD} & \textbf{STFPM} \\
    \hline
    {carpet} & 98.2/{\color{RoyalBlue}78.4} & 98.8/71.2 & 98.7/72.7 & 98.8/73.2 & 98.8/56.2 & {\color{red}99.2}/76.7 \\
    {grid} & 98.6/85.0 & 97.4/72.2 & 98.0/82.3 & 96.7/69.6 & {\color{red}99.3}/{\color{RoyalBlue}88.3} & 99.2/69.5 \\ 
    {leather} & 98.9/{\color{RoyalBlue}96.6} & 97.4/84.2 & 98.9/95.6 & 98.9/90.5 & 99.1/92.4 & {\color{red}99.6}/83.5 \\ 
    {tile} & 95.7/{\color{RoyalBlue}87.1} & 95.8/76.0 & 94.9/85.9 & 94.9/74.2 & 95.4/69.0 & {\color{red}97.1}/76.0 \\  
    {wood} & 90.8/84.9 & 95.0/82.0 & 93.2/84.0 & 93.9/84.5 & 94.9/84.9 & {\color{red}96.9}/{\color{RoyalBlue}85.2} \\  
    {bottle} & 97.8/{\color{RoyalBlue}92.3} & 98.5/59.3 & 98.0/84.4 & 98.3/92.2 & 98.3/76.4 & {\color{red}98.7}/59.9 \\ 
    {cable} & 93.8/78.2 & 95.6/68.3 & {\color{red}98.0}/84.3 & 97.2/{\color{RoyalBlue}89.0} & 96.4/53.9 & 94.9/73.8 \\ 
    {capsule} & 98.7/85.5 & 98.7/90.8 & {\color{red}98.8}/{\color{RoyalBlue}95.2} & 98.5/95.0 & 98.7/94.3 & 97.6/95.1 \\    
    {hazelnut} & 95.3/95.9 & 98.2/95.7 & 98.4/97.1 & 98.6/{\color{RoyalBlue}97.9} & 98.8/96.5 & {\color{red}99.1}/95.2 \\  
    
    {metal\_nut} & {\color{red}98.6}/82.7 & 97.4/76.1 & 98.5/84.4 & 96.1/{\color{RoyalBlue}86.5} & 97.0/82.4 & 98.2/81.8 \\   
    
    {pill} &97.5/85.3 & {\color{red}98.0}/90.7 & 97.5/{\color{RoyalBlue}94.6} & 95.2/92.7 & 97.4/91.2 & 95.8/88.0 \\
    
    {screw} & 98.1/85.0 & 97.4/93.9 & 99.2/95.7 & 98.7/94.8 & {\color{red}99.6}/{\color{RoyalBlue}97.0} & 98.9/93.6 \\
    
    {toothbrush} & 95.2/72.6 & 98.2/88.2 & 98.7/97.1 & {\color{red}99.0}/{\color{RoyalBlue}97.6} & 98.9/93.2 & {\color{red}99.0}/91.9\\
    
    {transistor} & 92.6/78.3 & 85.9/53.7 & 96.7/75.2 & {\color{red}97.6}/{\color{RoyalBlue}86.5} & 89.1/66.6 & 82.3/59.5 \\
    
    {zipper} & 95.9/74.3 & 96.3/70.7 & 98.1/86.6 & 97.2/{\color{RoyalBlue}88.0} & {\color{red}98.5}/78.0 & 98.1/78.6 \\
    \hline
    \textbf{Model AVG} & 96.4/84.1 & 96.6/78.2 & \textbf{{\color{red}97.7}}/\textbf{{\color{RoyalBlue}87.7}} & 97.3/87.5 & 97.3/81.4 & 97.0/80.6 \\
    \bottomrule
  \end{tabular}
  \end{adjustbox}
    \caption{MVTec pixel-level AUROC. Each cell carries the results for ImageNet/Fractals.}
        \label{tab:mvtec_pixel_auroc}
\end{table}

\begin{table}[htbp]
   \begin{adjustbox}{width=\columnwidth}
   \centering\fontsize{7.7}{11}\selectfont
  \begin{tabular}{l|cccccccc|}
    \toprule
    \textbf{Class} & \textbf{FastFlow} & \textbf{C-Flow} & \textbf{PatchCore} & \textbf{PaDiM} & \textbf{RD} & \textbf{STFPM} \\
    \hline
    {carpet} & --/51.3 & 93.8/33.1 & 92.7/31.4 & {\color{red}95.3}/39.6 & 94.8/24.8 & 97.0/{\color{RoyalBlue}51.9} \\
    {grid} & 95.1/63.2 & 90.8/40.3 & 90.1/60.7 & 89.0/41.1 & {\color{red}97.3}/{\color{RoyalBlue}70.2}& 97.0/31.6 \\
    
    {leather} & 98.3/{\color{RoyalBlue}89.8} & 90.8/47.9 & 96.3/76.7 & 98.0/68.9 & 97.9/69.0 & {\color{red}99.0}/51.6 \\
    
    {tile} & 87.4/{\color{RoyalBlue}72.1} & 90.2/63.3 & 79.6/69.0 & 86.3/64.3 & 87.5/45.1 & {\color{red}92.4}/49.5 \\
    
    {wood} & 89.3/{\color{RoyalBlue}75.0} & 88.6/50.7 & 84.6/54.9 & 91.6/65.5 & 91.3/70.3 & {\color{red}95.7}/62.7 \\
    
    {bottle} & 88.7/76.1 & 93.5/28.1 & 92.3/64.7 & 95.1/{\color{RoyalBlue}77.4} & 95.3/53.2 & {\color{red}96.2}/22.5 \\
    
    {cable} & 80.3/38.6 & 84.8/29.9 & {\color{red}91.1}/46.8 & 88.5/{\color{RoyalBlue}62.5} & 90.1/41.4 & 89.0/30.4 \\
    
    {capsule} & 92.4/59.3 & 91.0/73.9 & 92.3/75.1 & 91.1/77.6 & {\color{red}93.0}/81.8 & 91.1/{\color{RoyalBlue}81.9} \\
    
    {hazelnut} & 95.2/89.7 & 95.1/86.2 & 94.4/87.0 & 95.0/{\color{RoyalBlue}90.1} & 96.3/{\color{RoyalBlue}90.1} & {\color{red}97.6}/87.6 \\
    
    {metal\_nut} & 92.8/47.5 & 87.2/27.4 & 91.9/49.4 & 91.9/{\color{RoyalBlue}54.1} & 93.8/40.0 & {\color{red}95.4}/36.8 \\
    
    {pill} & 91.3/68.9 & 93.4/65.0 & 93.8/83.8 & 94.4/{\color{RoyalBlue}85.6} & {\color{red}96.2}/82.2 & 95.1/72.7 \\
    
    {screw} & 91.2/59.9 & 89.2/80.3 & 95.5/84.0 & 94.7/83.6 & {\color{red}97.7}/{\color{RoyalBlue}88.5} & 95.0/78.8\\
    
    {toothbrush} & 77.8/28.3 & 82.9/64.1 & 86.2/82.7 & {\color{red}93.2}/{\color{RoyalBlue}91.6} & 91.6/79.4 & 92.9/70.4 \\
    
    {transistor} & 79.1/44.4 & 73.8/21.8 & {\color{red}94.0}/42.3 & {\color{red}94.0}/{\color{RoyalBlue}62.4} & 79.2/41.1 & 69.4/16.0 \\
    
    {zipper} & 87.8/41.8 & 87.7/30.2 & 92.5/{\color{RoyalBlue}67.7} & 91.3/64.2 & {\color{red}95.3}/50.4 & 94.2/38.3 \\
    
    \hline
    \textbf{Model AVG} & 89.1/60.4 & 88.9/49.5 & 91.2/65.1 & 92.6/\textbf{{\color{RoyalBlue}68.6}} & \textbf{{\color{red}93.2}}/61.8 & 93.1/52.2 \\
    \bottomrule
  \end{tabular}
  \end{adjustbox}
    \caption{MVTec AUPRO. Each cell carries the results for ImageNet/Fractals.}
    \label{tab:mvtec_aupro}
\end{table}
In this section, we analyze in depth the experimental results of the chosen AD algorithms. Note that, except for CutPaste, none of the algorithms had their model weights fine-tuned. In each table the reported accuracies are expressed in percentage, the best result for each method is marked in red for ImageNet and blue for Fractals pre-training; in addition, each cell contains the results for ImageNet/Fractals. 

For MVTec the corresponding result table are \ref{tab:mvtec_image_auroc}, \ref{tab:mvtec_pixel_auroc} and \ref{tab:mvtec_aupro}. In \cref{tab:mvtec_image_auroc} we observe that PatchCore is the winning approach followed by RD when using ImageNet as they both solve 7 of the 15 sub-datasets. With Fractals CutPaste solves 7 of the 15 classes achieving the highest average image-level AUROC of 80.9\%. For some classes Fractals surpass the performance of ImageNet: \textit{grid} when using CutPaste and PANDA, \textit{wood} with FastFlow and \textit{toothbrush} with C-Flow, PaDiM, RD and CutPaste. In \cref{tab:mvtec_pixel_auroc} we can see that PatchCore reaches the heights pixel-level AUROC for both ImageNet and Fractals, followed by PaDiM. When using the AUPRO, Fractals performance drops. As shown in \cref{tab:mvtec_aupro} C-Flow is the methods that have the biggest drops in localization performance when compared with the results in \cref{tab:mvtec_pixel_auroc}. PaDiM reaches the highest AUPRO of 68.9\%. Note that the AUPRO metric with the carpet class for the FastFlow pre-trained with ImageNet is missing. Anomalib (\cite{akcay2022anomalib}), the repository used for the evaluation, led to a value of 1.21, which is a bug, thus, we did not report any value.

For VisA the corresponding result tables are \ref{tab:visa_image_auroc},  \ref{tab:visa_pixel_auroc} and  \ref{tab:visa_aupro}. As shown in  \cref{tab:visa_image_auroc} when using Fractals, PatchCore reached the best accuracy of 80.4\% followed by CutPaste with 79.2\%. We have some cases where Fractals surpass ImageNet results: \textit{capsules} with PatchCore and PANDA, \textit{macaroni2} with CutPaste and PANDA, \textit{pcb1} PaDiM and CutPaste and for \textit{pcb2} with PatchCore, PaDiM and CutPaste. \cref{tab:visa_pixel_auroc} shows the pixel-level AUROC. For ImageNet the best approach is RD for Fractals PatchCore. On average the pixel-level performance differs around 11\% between ImageNet and Fractals. Here, too, using AUPRO metrics results in a performance drop as shown in \cref{tab:visa_aupro}.

Overall for both datasets is clear that memory-based methods seem the more suitable when using Fractals, while flow-based methods are the ones with the lowest performance. CutPaste works well with fractals reaching the first position on MVTec and the second on VisA. For ImageNet the best results remain between teacher-student and memory-based methods. When using AUROC metrics, using fractal images leads to promising results both at the image and pixel level. The performance drops when using AUPRO, indicating that small defects are not well localized. Meanwhile, ImageNet weights can maintain good performance across all metrics.

\begin{table*}[htbp]
   \centering\fontsize{7.7}{11}\selectfont
  \begin{tabular}{l|cccccccc}
    \toprule
    \textbf{Class} & \textbf{FastFlow} & \textbf{C-Flow} & \textbf{PatchCore} & \textbf{PaDiM} & \textbf{RD} & \textbf{STFPM} & \textbf{CutPaste} & \textbf{PANDA} \\
    \hline
    {candle} & 94.2/69.7 & 92.2/69.1 &{\color{red} 97.9}/{\color{RoyalBlue}83.1} & 92.6/79.7 & 94.0/76.2 & 80.7/70.7 & 96.6/77.9 & 88.4/67.9 \\
    {capsules} & 85.6/49.8 & 79.4/69.1 & 68.4/{\color{RoyalBlue}79.6} & 65.6/62.7 & 84.6/62.7 & {\color{red}88.4}/68.4 & 83.7/71.4 & 57.1/68.2 \\  
   {cashew} & 89.0/90.9 & 91.9/78.6 & 95.6/{\color{RoyalBlue}91.8} & 88.1/82.3 & {\color{red}96.3}/65.0 & 86.1/80.2 & 82.7/73.1 & 91.6/90.2 \\    
    {chewinggum} & 95.8/{\color{RoyalBlue}91.6} & 98.4/80.1 & {\color{red}99.4}/81.9 & 98.3/71.7 & {\color{red}99.4}/67.8 & 98.2/73.5 & 96.6/86.0 & 92.2/69.0 \\  
    {fryum} & 78.0/61.1 & 78.0/71.4 & 91.6/{\color{RoyalBlue}82.6} & 84.6/80.7 & {\color{red}91.9}/70.8 & 89.2/60.7 & 93.4/75.8 & 84.5/74.8 \\
    
    {macaroni1} & 95.0/{\color{RoyalBlue}84.8} & 87.7/66.2 & 89.7/75.9 & 81.1/71.5 & {\color{red}96.3}/73.1 & 92.2/72.9 & 85.1/67.1 & 77.2/68.0 \\
    
    {macaroni2} & {\color{red}86.9}/52.4 & 76.8/58.0 & 71.7/59.6 & 62.0/60.8 & 80.8/62.7 & 84.3/59.1 & 63.1/{\color{RoyalBlue}75.5} & 58.7/67.3 \\
    
   {pcb1} & 95.2/72.4 & 90.9/54.6 & 95.1/89.8 & 83.2/83.3 & {\color{red}97.0}/62.9 & 87.6/36.0 & 89.4/{\color{RoyalBlue}92.7} & 87.0/59.5 \\
    
    {pcb2} & {\color{red}95.2}/80.7 & 80.0/29.8 & 93.5/94.7 & 82.7/88.3 & 96.8/85.6 & 90.3/30.2 & 93.6/{\color{RoyalBlue}95.5} & 91.3/83.7 \\
    
    pcb3& 94.4/50.5 & 85.6/56.6 & 91.9/71.1 & 78.9/76.5 & {\color{red}96.5}/{\color{RoyalBlue}93.2} & 90.0/64.0 & 89.7/72.6 & 78.1/64.3 \\
    
    {pcb4} & 97.0/69.8 & 97.1/83.9 & 99.5/90.6 & 93.2/94.0 & {\color{red}99.8}/{\color{RoyalBlue}96.5} & 95.5/81.4 & 97.4/95.0 & 96.5/83.0 \\
    
    {pipe\_fryum} & {\color{red}99.5}/64.8 & 94.8/64.5 & 98.5/64.4 & 96.7/66.1 & 97.3/{\color{RoyalBlue}74.6} & 92.6/64.3 & 76.3/67.3 & 80.1/59.8 \\
    \hline
    \textbf{Model AVG} & 92.1/69.9 & 87.7/65.2 & 91.1/\textbf{{\color{RoyalBlue}80.4}} & 83.9/76.5 & \textbf{{\color{red}94.2}}/74.3 & 89.6/63.4 & 87.3/79.2 & 81.9/71.3 \\
    \bottomrule
  \end{tabular}
  \caption{VisA image-level AUROC. Each cell carries the results for ImageNet/Fractals.}
  \label{tab:visa_image_auroc}
\end{table*}

\begin{table}[htbp]
   \begin{adjustbox}{width=\columnwidth}
   \centering\fontsize{7.7}{11}\selectfont
  \begin{tabular}{l|cccccccc}
    \toprule
    \textbf{Class} & \textbf{FastFlow} & \textbf{C-Flow} & \textbf{PatchCore} & \textbf{PaDiM} & \textbf{RD} & \textbf{STFPM} \\
    \hline
    {candle} & {\color{red}99.2}/80.7 & 98.7/74.6 & 98.9/82.6 & 98.7/77.4 & 99.0/85.9 & 98.9/{\color{RoyalBlue}86.5} \\
    
    {capsules} & 98.2/84.2 & 97.0/82.2 & 97.6/90.9 & 96.3/90.2 & {\color{red}99.6}/{\color{RoyalBlue}92.5} & 99.3/76.8 \\
    
    {cashew} & 98.2/89.6 & {\color{red}99.1}/91.8 & 99.0/75.1 & 98.6/74.3 & 95.1/41.4 & 97.0/{\color{RoyalBlue}92.8} \\
    
    {chewinggum} & {\color{red}99.2}/{\color{RoyalBlue}96.9} & 98.8/94.1 & 98.9/87.3 & 98.9/69.1 & 98.7/86.7 & 99.1/93.3 \\
    
    {fryum} & 89.0/88.5 & {\color{red}96.5}/89.0 & 94.9/{\color{RoyalBlue}94.2} & 95.5/94.1 & 96.3/92.1 & 95.4/87.0 \\
    
    {macaroni1} & 96.3/98.0 & 98.6/91.3 & 98.2/95.2 & 97.4/93.8 & {\color{red}99.5}/{\color{RoyalBlue}98.6} & 99.4/97.3 \\
    
    {macaroni2} & 98.7/94.9 & 97.5/90.9 & 96.9/91.8 & 94.9/91.0 & 99.2/{\color{RoyalBlue}96.2} & {\color{red}99.6}/95.5 \\
    
    {pcb1} & {\color{red}99.7}/94.0 & 99.1/87.2 & 99.5/{\color{RoyalBlue}98.4} & 98.7/89.6 & 99.6/31.1 & 99.4/47.7 \\
    
    {pcb2} & {\color{red}98.7}/91.0 & 96.1/84.0 & 97.8/92.8 & 97.3/{\color{RoyalBlue}94.3} & 98.5/89.5 & 97.3/76.8 \\
    
    {pcb3} & 93.5/85.4 & 97.3/86.2 & 98.2/92.7 & 97.2/{\color{RoyalBlue}96.1} & {\color{red}99.0}/95.0 & 98.1/89.3 \\
    
    {pcb4} & {\color{red}98.4}/77.0 & 97.8/81.9 & 97.7/83.2 & 96.5/88.4 & 98.1/{\color{RoyalBlue}94.3} & 98.2/89.6 \\
    
    {pipe\_fryum} & 98.3/90.7 & 98.6/95.8 & 98.8/96.0 & {\color{red}98.9}/96.9 & 98.7/{\color{RoyalBlue}97.2} & 97.9/96.7 \\
    \hline
    \textbf{Model AVG} & 97.3/89.2 & 97.9/87.4 & 98.0/\textbf{{\color{RoyalBlue}90.0}} & 97.4/87.9 & \textbf{{\color{red}98.4}}/83.4 & 98.3/85.8 \\
    \bottomrule
  \end{tabular}
  \end{adjustbox}
    \caption{VisA pixel-level AUROC. Each cell carries the results for ImageNet/Fractals.}
  \label{tab:visa_pixel_auroc}
\end{table}


\begin{table}[htbp]
   \begin{adjustbox}{width=\columnwidth}
   \centering\fontsize{7.7}{11}\selectfont
  \begin{tabular}{l|cccccccc}
    \toprule
    \textbf{Class} & \textbf{FastFlow} & \textbf{C-Flow} & \textbf{PatchCore} & \textbf{PaDiM} & \textbf{RD} & \textbf{STFPM} \\
    \hline
    {candle} & {\color{red}94.8}/42.5 & 92.7/43.2 & 94.3/{\color{RoyalBlue}72.8} & 94.0/49.4 & 94.1/71.4 & 94.5/61.8 \\
    {capsules} & 90.6/45.9 & 75.3/51.3 & 67.8/{\color{RoyalBlue}61.9} & 68.7/56.8 & 93.1/51.7 & {\color{red}95.3}/44.6 \\
    cashew & 81.1/{\color{RoyalBlue}81.3} & {\color{red}92.5}/74.3 & 89.4/42.6 & 84.6/37.7 & 87.4/38.1 & 92.1/77.0 \\
    chewinggum & 84.4/62.7 & {\color{red}88.9}/53.7 & 84.7/43.0 & 86.5/29.8 & 80.5/48.0 & 83.0/{\color{RoyalBlue}68.6} \\
    {fryum} & 69.7/68.7 & 81.0/69.7 & 80.2/72.2 & 70.1/70.6 & {\color{red}88.4}/{\color{RoyalBlue}77.8} & 85.9/65.3 \\
    {macaroni1} & 87.1/{\color{RoyalBlue}95.1} & 90.7/79.1 & 91.8/81.8 & 87.6/67.3 & {\color{red}95.0}/87.3 & 94.8/88.0 \\
    {macaroni2} & 93.9/69.4 & 83.4/60.9 & 86.9/58.3 & 71.5/54.9 & 92.7/75.4 & {\color{red}95.5}/{\color{RoyalBlue}76.2} \\
    {pcb1} & 92.5/64.9 & 88.1/49.7 & 89.9/{\color{RoyalBlue}77.8} & 87.5/74.4 & {\color{red}95.6}/18.0 & 92.3/14.4 \\
    {pcb2} & 85.7/68.5 & 76.7/54.4 & 83.7/{\color{RoyalBlue}78.9} & 77.6/78.8 & {\color{red}90.4}/67.2 & 85.3/33.7 \\
    {pcb3} & 79.6/42.1 & 73.5/64.9 & 80.4/78.5 & 70.6/80.7 & {\color{red}91.0}/{\color{RoyalBlue}88.4} & 89.6/77.1 \\
    {pcb4} & 89.0/30.6 & 86.2/42.8 & 84.6/44.1 & 79.1/52.6 & 88.1/{\color{RoyalBlue}75.7} & {\color{red}89.7}/66.1 \\
    {pipe\_fryum} & 86.1/78.0 & 92.9/87.0 & 93.4/78.5 & 90.5/79.2 & {\color{red}95.0}/{\color{RoyalBlue}88.9} & 93.7/{\color{RoyalBlue}88.9} \\
    \hline
    \textbf{Model AVG} & 86.2/62.5 & 85.2/60.9 & 85.6/65.9 & 80.7/61.0 & 90.9/\textbf{{\color{RoyalBlue}65.7}} & \textbf{{\color{red}91.0}}/63.5 \\
    \bottomrule
  \end{tabular}
  \end{adjustbox}
  \caption{VisA AUPRO. Each cell carries the results for ImageNet/Fractals.}
  \label{tab:visa_aupro}
\end{table}

\subsection{Learned weights}
\begin{figure*}[h!]
    \centering
    \includegraphics[width=0.8\textwidth]{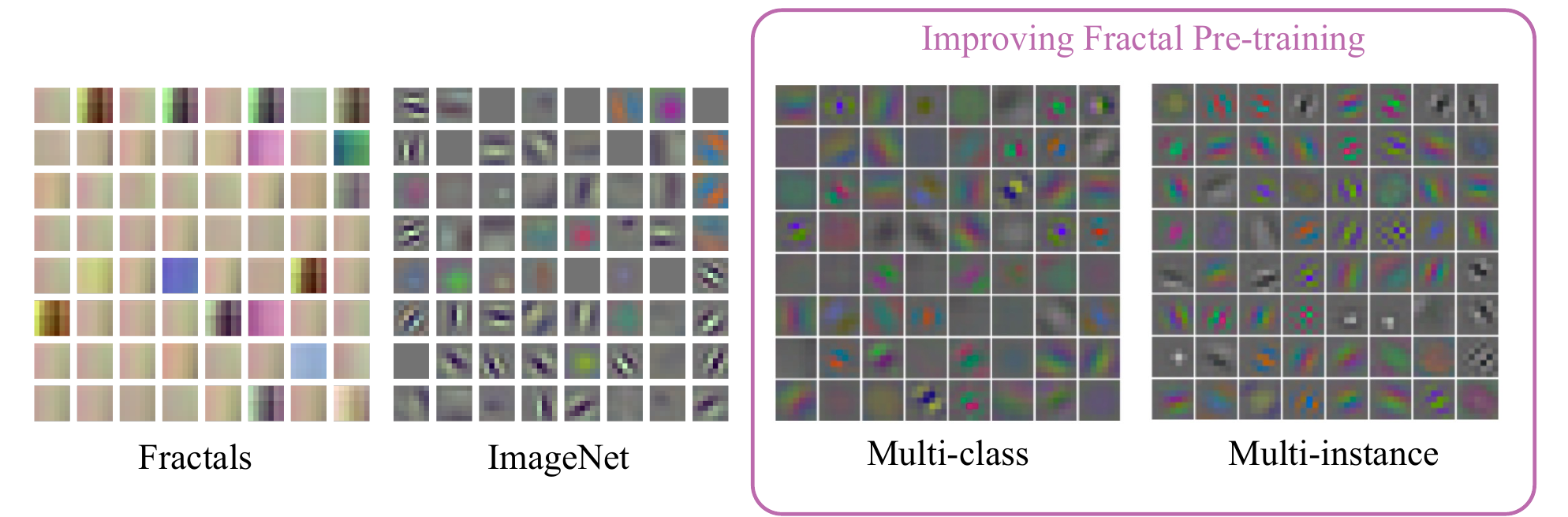}
    \caption{Comparison of the filters from the first layer of ResNet18 pre-trained with Fractals (left) and ImageNet (right). In the purple box, we can find two images taken from \cite{anderson2022improving} showing the first layer of ResNet50 learned using fractals with multi-class and multi-instance prediction.}
    \label{fig:weights_first_layer}  
\end{figure*}

\cref{fig:weights_first_layer} shows the filters from the first layer of ResNet pre-trained using different methods.  We can see that the weight learned by us (Fractals) shows simple patterns, such as solid vertical or horizontal lines. This could mean that the model has learned more basic features in the input data. In the purple box, we show the weights visualization taken from the original paper (\cite{anderson2022improving}) with the model trained with multi-class and multi-instance. Multi-instance is a more advanced training where there are multiple classes per image. Multi-instance prediction learns first-layer filters that are very similar to those learned from ImageNet pre-training and also their multi-class weights show more complex patterns meaning their model has likely learned to detect more intricate and nuanced features in the input data. 

\subsection{Comparison between object categories}
\begin{figure}[h!]
    \centering
    \begin{subfigure}{0.45\textwidth}
        \centering
        \includegraphics[width=\linewidth]{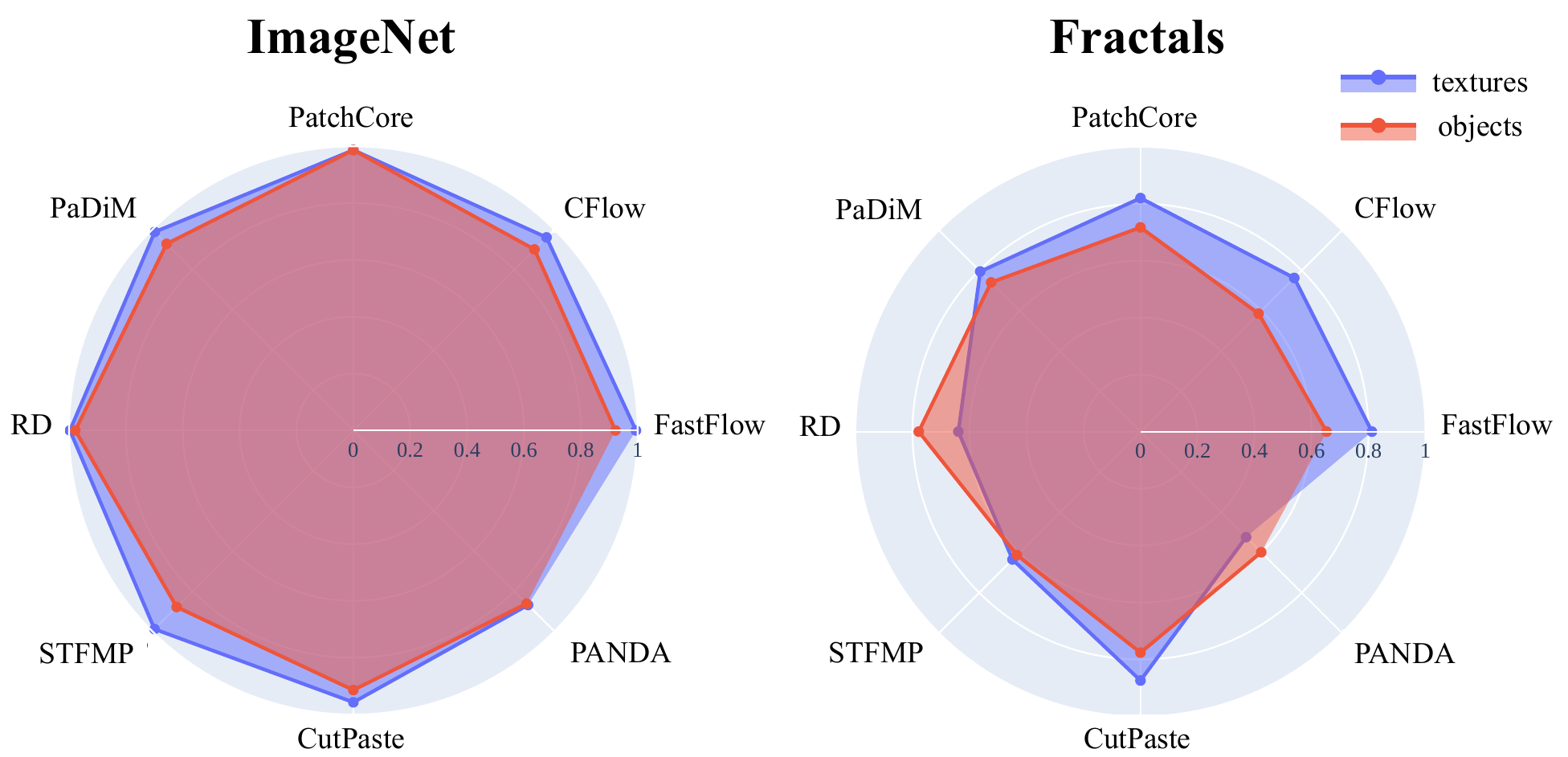}
        \caption{MVTec}
        \label{fig:spider_mvtec}
    \end{subfigure}
    \begin{subfigure}{0.45\textwidth}
        \centering
        \includegraphics[width=\linewidth]{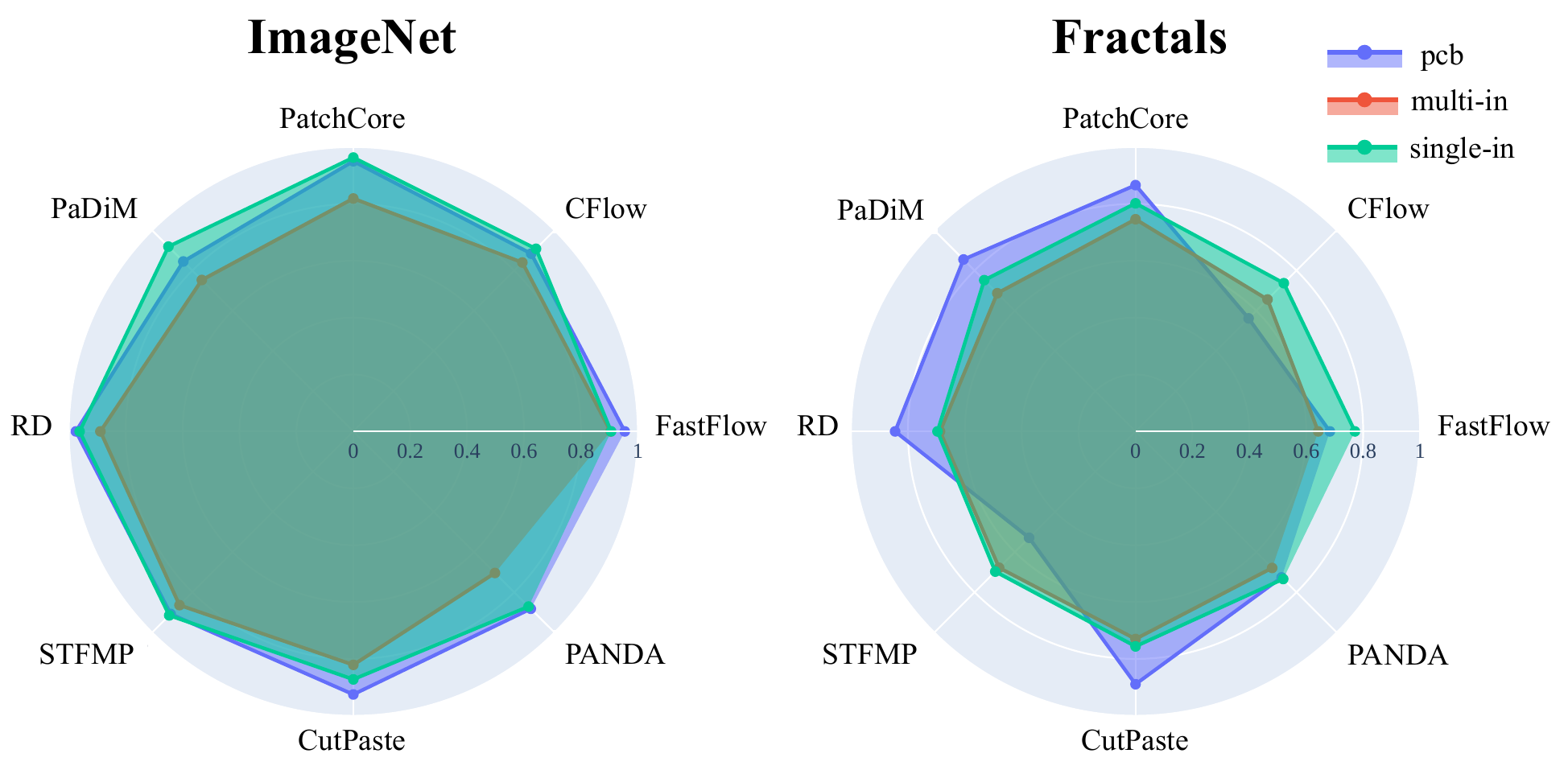}
        \caption{VisA}
        \label{fig:spider_VisA}
    \end{subfigure}
    \caption{Spider chart representing average image-level AUROC grouping MVTec Ad and VisA classes into different object categories.}
    \label{fig:spider_plot}
\end{figure}

For MVTec the 15 sub-dataset can be divided into \textit{textures} (carpet, grid, leather, tile, wood) and \textit{objects} (bottle, cable, capsule, hazelnut, metal\_nut, pill, screw, toothbrush, transistor, zipper). For VisA the 12 sub-classes are divided into \textit{pcb} (pcb1, pcb2, pcb3, pcb4), images with multi-instance in a view \textit{multi-in} (candle, capsules, macaroni1, macaroni2) and image with single-instance in a view \textit{single-in} (cashew, chewinggum, fryum, pipe\_fryum). In \cref{fig:spider_plot} we can see a qualitative visualization of the image-level AUROC accuracy group by object categories. Focusing on \cref{fig:spider_mvtec} ImageNet leads to good performance for both \textit{textures} in blue and \textit{objects} in red for all the methods. The larger blue area shows a higher performance for the \textit{texture} category. Also with Fractals, we have the same behaviour except for RD and PANDA with \textit{objects} having respectively +1.9\% and +0.7\% compared to \textit{textures}. The bigger difference between \textit{textures} and \textit{objects} can be seen for flow-based methods with +7.2\% and +6\% for FastFlow and C-Flow. \cref{fig:spider_VisA} shows the results for VisA where it is clear that for both ImageNet and Fractals all the methods underperform for \textit{multi-in}. Our intuition is that this behaviour is more method-related rather than weight-related. The proposed methods are specialised to perform well on MVTec which is composed of images with single objects in a view. For ImageNet \textit{pcb} and \textit{sinle-in} have comparable performance, while for Fractals the results are quite variable. Overall is clear that ImageNet is the winning dataset, however, Fractals' results are quite promising considering that we are training on completely abstract images without any fine-tuning.

\label{sec:object_categories}
\subsection{Impact of feature hierarchy}
\label{sec:features_hierarchy}
\begin{figure}[h!]
    \centering
    \includegraphics[width=0.5\textwidth]{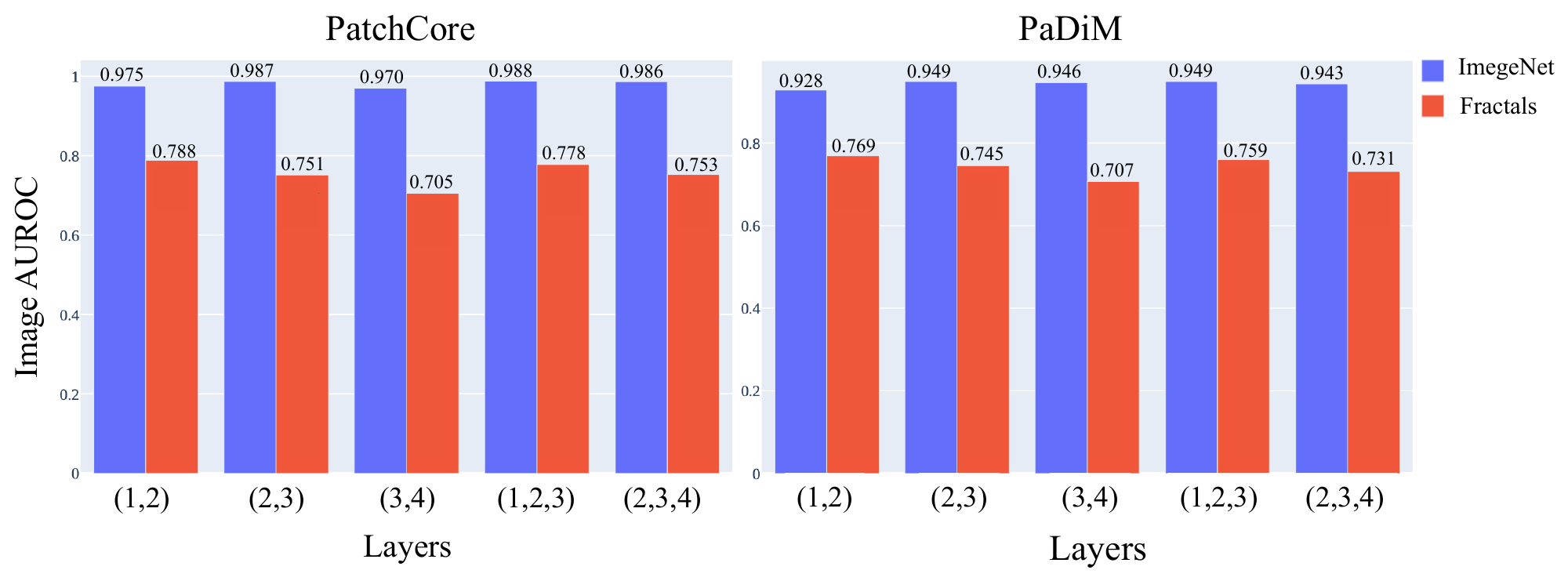}
    \caption{Comparison between ImageNet (blue) and Fractals (red) of the average image-level AUROC when using different feature hierarchies.}
    \label{fig:image_auroc}
\end{figure}

\begin{figure}[h!]
    \centering
    \includegraphics[width=0.5\textwidth]{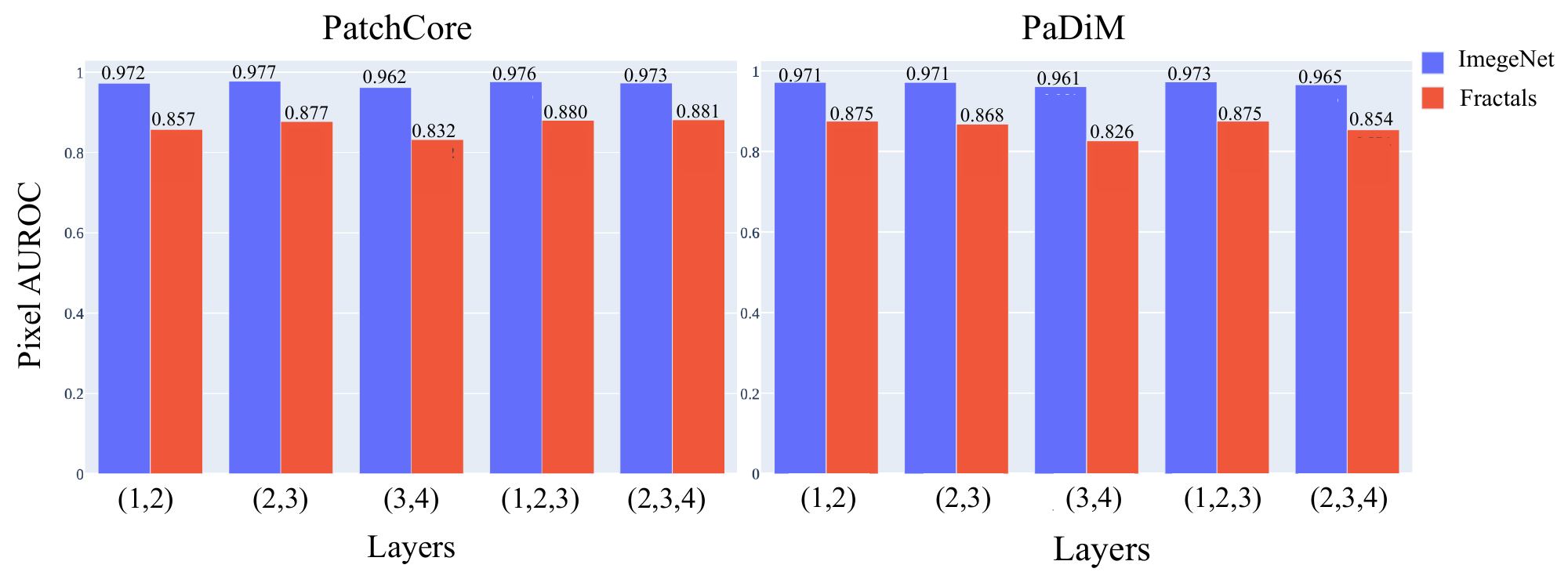}
    \caption{Comparison between ImageNet (blue) and Fractals (red) of the average pixel-level AUROC when using different feature hierarchies.}
    \label{fig:layers_pixel_auroc}
\end{figure}

\begin{figure}[h!]
    \centering
    \includegraphics[width=0.5\textwidth]{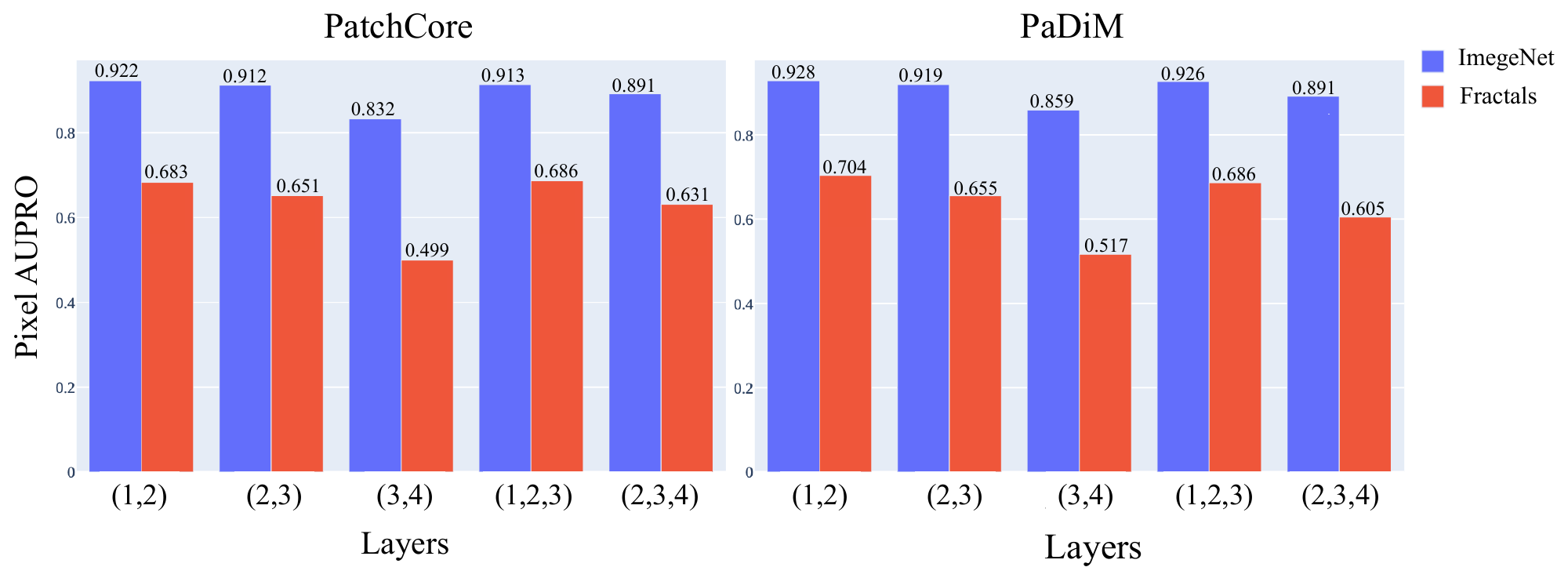}
    \caption{Comparison between ImageNet (blue) and Fractals (red) of the average AUPRO when using different feature hierarchies.}
    \label{fig:layers_aupro}
\end{figure}
Feature maps from ResNet-like architectures, which play an important role, can be divided into hierarchy-level $j \in \{1,2,3,4\}$. For example, using the last level for feature representation introduces two problems (i) the loss of more localized nominal information, as the last layers of the network extract more high-level features, (ii) and feature bias towards the task of natural image classification which has only little overlap with industrial anomaly detection (\cite{roth2022towards}). As pointed out by \cite{kataoka2020pre} and \cite{anderson2022improving}, models pre-trained on fractal images are unbiased when compared to ImageNet, so we studied the impact of features hierarchy when using Fractals. We use PatchCore and PaDiM which rely on $j \in \{2,3\}$ and $j \in \{1,2,3\}$ for feature representation. \cref{fig:image_auroc} shows the average image-level accuracy on MVTec when considering different $j$. Focusing on ImageNet (blue), the results with different hierarchies are quite stable for both methods. Nevertheless, there is not a huge bust in performance when combining three hierarchies instead of two for both ImageNet and Fractals. This outcome is significant as memory-based methods necessitate a large amount of memory during initialization, which increases with the number of features involved. The pixel-level AUROC is reported in \cref{fig:layers_pixel_auroc}; for both ImageNet and Fractals there is a clear drop in performance with $j \in \{3,4\}$. When using Fractals the best results are obtained with $j \in \{2,3\}$ for PatchCore and $j \in \{1,2\}$ and $j \in \{1,2,3\}$ for PaDiM. We can see that the difference between the results using ImageNet and the results using Fractals is relatively small. This difference increases when considering the AUPRO metrics, see \cref{fig:layers_aupro}. When using layers $j \in \{3,4\}$ with Fractals we reach an accuracy of 49.9\% for PatchCore and 51.7\% for PaDiM. 

For all the cases the best performance is obtained when using low-level features $j \in \{1,2\}$ meaning that low-level features from fractals can help more than high-level features in solving the AD task. This could be related to the fact that fractal structures cover more real-world patterns than ImageNet (\cite{yamada2022point}) our intuition is that low-level features capture these simpler patterns, that can be found in nature, rather than high-level features which are correlated more to the complex geometric structure of the attractor $\mathcal{A}_S$. 

\subsection{Qualitative results}
\begin{figure*}[h!]
    \centering
    \includegraphics[width=0.71\textwidth]{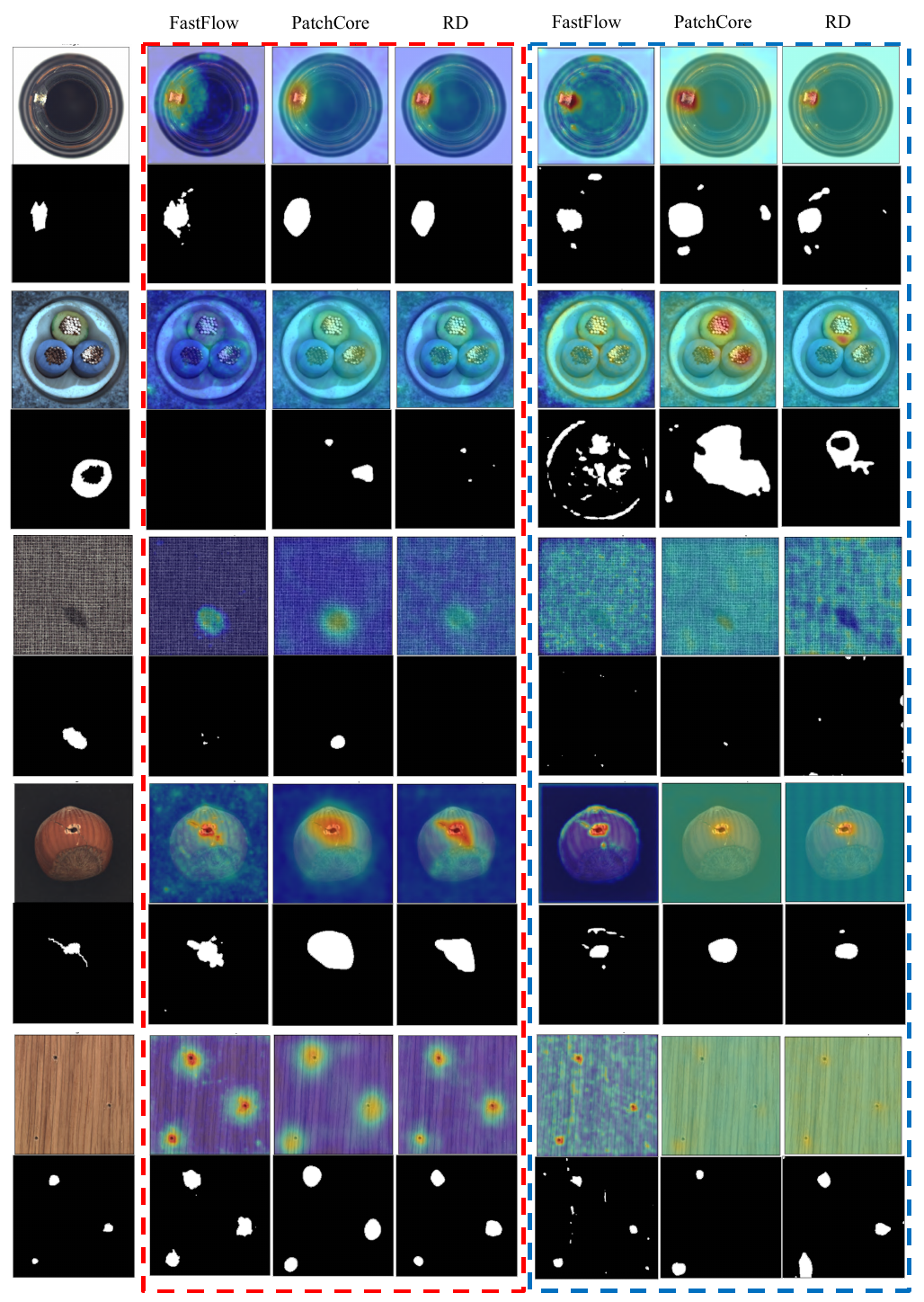}
    \caption{Qualitative visualization for the MVTec's classes: \textit{bottle}, \textit{cable}, \textit{carpet}, \textit{hazelnut} and \textit{wood}. In the first column, we have the original image and the ground-truth. In the \textcolor{red}{red} box we have the anomaly score and predicted segmentation mask for ImageNet pre-training and in the \textcolor{blue}{blue} box for Fractals.}
    \label{fig:qualitative_mvtec}
\end{figure*}

In \cref{fig:qualitative_mvtec} we can see some qualitative results on MVTec's classes: \textit{bottle}, \textit{cable}, \textit{carpet}, \textit{hazelnut} and \textit{wood}. In the red box, we have the anomaly score and predicted segmentation mask for ImageNet pre-training and in the blue box for Fractals. It is interesting to notice that for \textit{cable} the anomaly type is called \textit{cable\_swap} so rather than a structural defect such as scratches, dents, colour spots or cracks, we are facing a misplacement, a violation of the position of an object which can be seen as a logical anomaly. We can see from the figure that none of the methods both using ImageNet or Fractals can predict the correct segmentation mask. We also observe that Fractals tend to fail when localizing anomalies with low contrast with the background like for \textit{carpet}.

\section{Discussion}
We chose to focus on industrial defect detection because it is a specialized area that is limited by the scarcity of data, mainly due to stringent privacy regulations and the high cost of labelling. Large datasets have the potential to violate ethical or privacy standards. In such cases, these datasets may be suspended from functionality, face ownership shifts, or be abruptly withdrawn. Such scenarios can complicate the tracking of lineages, raise concerns regarding privacy and data integrity, and make it difficult to assign credit to data sources. In industrial defect detection with limited access to quality labelled data, relying solely on supervised methods becomes increasingly impractical due to the data-hungry nature of deep learning models. Our work deviates from existing literature by focusing on utilizing abstract images, such as fractals, for pre-training without altering existing methodologies. Our study is a preliminary examination of how generated data impacts established methods, shedding light on the potential of fractals in anomaly detection. While conventional approaches rely on fractals transfer learning ability, we deliberately omit this step to maintain consistency with the original methods (only CutPaste includes fine-tuning by default) showcasing the effectiveness of fractals in anomaly detection.

\section{Conclusions}
This paper investigated the potential utility of using abstract, computer-generated fractal images to pre-train feature extractors in unsupervised visual anomaly detection systems. We conducted a systematic analysis of 8 state-of-the-art AD methods and tested their performance on 27 object classes each having different types of anomalies. Experiments reveal that memory-based methods and CutPaste seem statistically better than others and their results vary depending on the type of objects' class, emphasizing the importance of anomaly type selection when considering fractal images. Although pre-training with ImageNet remains a clear winner on this task, the fact that we were able to achieve relatively good performance by learning weight from completely abstract images is quite stunning. 

In future work, our studies may be continued in a variety of ways. First, as the learned weights exhibit simple patterns, such as solid vertical or horizontal lines,  multi-instance training should be taken into consideration since it has proven to learn weights more similar to ImageNet obtaining models that better generalize to the downstream tasks. Second, we observe that in the literature little effort has been put into synthesizing abnormal samples via data augmentation which is a difficult but important task. More attention should be given to self-supervised methods like CutPaste since they involve fine-tuning, in line with the fractals literature. Third, exploring fractals' performance under few-shot learning should be investigated. Fractal pre-trained weights could reduce data needed for fine-tuning benefiting fields affected by limited data like medicine.